\documentclass[10pt,twocolumn,letterpaper]{article}
\usepackage{graphicx}
\usepackage{longtable}
\usepackage{xcolor}
\usepackage[table]{xcolor}
\usepackage{placeins}
\usepackage{csquotes}
\usepackage[pagenumbers]{wacv} 

\definecolor{wacvblue}{rgb}{0.21,0.49,0.74}
\usepackage[pagebackref,breaklinks,colorlinks,allcolors=wacvblue]{hyperref}

\def\wacvPaperID{2286} 
\def\confName{WACV}
\def\confYear{2026}

\title{GeoThought: A Dataset for Enhancing Mathematical Geometry Reasoning in Vision-Language Models}

\author{Nannan Shi\textsuperscript{\rm 1}\thanks{\ \  This work was done at Baidu Inc.},\ 
Chuanyu Qin\textsuperscript{\rm 2},\ 
Shipeng Song\textsuperscript{\rm 1},\ 
Man Luo\textsuperscript{\rm 3}\thanks{\ \  Corresponding author: Man Luo.} \\
\textsuperscript{\rm 1}Baidu Inc., Beijing, China \\
\textsuperscript{\rm 2}Institute of Information Engineering, Chinese Academy of Sciences, Beijing, China \\
\textsuperscript{\rm 3}Intel Lab, Intel, Santa Clara, USA \\
  \texttt{shinannan321@gmail.com, qinchuanyu24@mails.ucas.ac.cn} \\
  \texttt{songshipeng@baidu.com, man.luo@intel.com} \\
}

\begin{document}
\maketitle
\begin{abstract}
Large language models (LLMs) have demonstrated strong reasoning capabilities in text-based mathematical problem solving; however, when adapted to visual reasoning tasks, particularly geometric problem solving, their performance substantially declines because geometric problems present unique challenges. Specifically, these challenges stem from two key factors: first, the intrinsic complexity of geometry requiring detailed image comprehension and multi-step reasoning, and second, the limitations of existing datasets which lack sufficient scale, diversity, and explicit reasoning traces, consequently hindering effective model training. To address these challenges, 
we developed the GeoThoughts dataset, a comprehensive geometric reasoning corpus with two subsets: Geo-Thought-6K with 6,243 samples and its augmented version Geo-Thought-Augmented-10K containing 10,834 samples. Each entry includes visual descriptions, step-by-step solutions, explicit reasoning chains, reflection steps, and final answers. Using this dataset, we developed GeoThought-MLLM, a mathematical reasoning multimodal model that generates detailed thinking processes during problem-solving. 
Our model outperforms existing benchmarks in geometric tasks, demonstrating that training with our Chain-of-Thought dataset improves geometric reasoning capabilities across both in-domain and out-of-domain settings.\footnote{Our code and dataset are available at: \url{https://github.com/xinlingdedeng/GeoThought}} Finally, we analyze failure cases and observe that errors primarily arise from incorrect interpretation of mathematical concepts or spatial misjudgment. By invoking CoT to correct these mistakes, the model produces correct answers.



\end{abstract}    
\section{Introduction}
\label{sec:intro}

Geometric reasoning represents one of the most challenging cognitive tasks in artificial intelligence, demanding models to simultaneously possess precise visual perception capabilities and sophisticated logical reasoning abilities\cite{klette2004digital}. The importance of geometric reasoning capabilities is evident across numerous domains, including mathematical education, engineering design, and scientific research. With the rapid advancement of Multimodal Large Language Models (MLLMs), extending powerful reasoning capabilities to geometric problem-solving that incorporates visual information has emerged as a critical research direction.

However, current state-of-the-art multimodal models still face significant challenges in geometric reasoning tasks. Although reasoning models such as OpenAI's o3~\cite{openai2025o3mini} and Deepseek-R1~\cite{guo2025deepseek} have demonstrated exceptional performance on pure text-based reasoning tasks, effectively transferring such reasoning capabilities to multimodal geometric domains has yielded limited success~\cite{liu2025othink}. More concerning is the observed degradation of reasoning capabilities during supervised fine-tuning processes~\cite{chen2025sft}. Recent evaluation studies ~\cite{wang2025benchmarking,xu2025geosense,xiang2025seephys}reveal that even top-tier models fall far short of human-level performance on complex geometric reasoning tasks, indicating substantial room for improvement in this critical domain.

The fundamental cause of this performance bottleneck lies in the severe lack of high-quality geometric reasoning training data. Geometric problems possess inherent complexity: they require models to accurately understand spatial relationships within geometric figures, identify key geometric elements, and execute multi-step logical deductions~\cite{chen2021geoqa,lu2021inter}. However, existing geometric datasets primarily focus on simple image-text description pairs, question-answer pairs, and final solutions~\cite{gao2023g,cao2022augmented,chen2022unigeo}, severely lacking detailed reasoning process data. This data scarcity constrains model development at two levels: first, the scale and diversity of existing datasets are insufficient to support large-scale model training; second, the absence of intermediate reasoning step annotations makes it difficult for models to learn human-like problem-solving strategies.

Chain-of-Thought (CoT) reasoning holds crucial value in addressing this predicament. CoT not only significantly improves model problem-solving accuracy by decomposing complex problems into a series of intermediate reasoning steps~\cite{wei2022chain}, but more importantly, makes the reasoning process more aligned with human cognitive patterns~\cite{feng2024far}. In the specific context of geometric reasoning, CoT enables models to systematically analyze geometric relationships, form hypotheses, execute recursive decomposition, and perform self-verification. Recent research demonstrates that multimodal models with CoT generation capabilities exhibit superior cognitive advantages in geometric problems~\cite{deng2024r}, further confirming the unique value of CoT in geometric reasoning.

To overcome existing technological bottlenecks, we designed an innovative data generation pipeline to construct high-quality geometric CoT datasets. Specifically, we first leverage problems and images from existing datasets to query teacher models for generating CoT sequences comprising complete reasoning processes and final answers. Subsequently, we use these problem-image pairs as seeds to synthesize five additional problems through prompt engineering techniques. For each generated problem, we query the teacher model eight times and retain only instances with 100\% consensus to ensure data reliability.

Based on the constructed GeoThoughts dataset, we trained GeoThought MLLM, a multimodal large language model capable of generating detailed Chain-of-Thought reasoning processes when solving geometric problems. Extensive experiments demonstrate substantial improvements: InternVL3-8B achieves performance improvements from 51.19\% to 73.21\% on GeoQA benchmarks when trained with our GeoThought-Augmented-10K dataset. These consistent gains across different model architectures conclusively demonstrate the superior cognitive advantages of Chain-of-Thought generation in geometric reasoning.
Our contributions are as follows:

\begin{itemize}
\item We developed a  data generation pipeline. This approach produces data that not only significantly enhances a model's geometric reasoning capabilities but also features a transferable methodology.
\item Leveraging the generated data, we trained GeoThought MLLMs for geometric reasoning, achieving substantial performance improvements over the same base models trained on other datasets. Our models surpass existing open-source state-of-the-art MLLMs and reach approximately ~94\% of the performance of much larger closed-source models.
\item We conducted an in-depth analysis of GeoThought MLLM's performance, paying special attention to cases of reasoning errors, and discussed potential solutions.
\end{itemize}

\section{Related Work}
\label{sec:related}
\textbf{Multi-modal Geometry Dataset.} Geometric problem solving serves as a crucial benchmark for evaluating mathematical reasoning capabilities in multimodal models. Early efforts in dataset construction relied primarily on manual annotation ~\cite{seo2015solving,sachan2017from,alvin2017synthesis}, establishing foundational resources but with limited scale. The field has since evolved through data augmentation approaches, yielding datasets like Geometry3K ~\cite{lu2021inter}, GeoQA ~\cite{chen2021geoqa}, UniGeo ~\cite{chen2022unigeo}, and the larger-scale Geo170K ~\cite{gao2023gllava}. Recent developments have further expanded scope and complexity: MATH-Vision ~\cite{wang2024measuring} introduced competition-level problems, MathV360K ~\cite{shi2024math} demonstrated synthetic data generation at scale, and GeoSense ~\cite{xu2025geosense} pioneered geometric principle annotation.

These geometric datasets often contain image descriptions, question-answer pairs about the images, and problem-solving steps. However, they frequently lack explicit thought chains, which are crucial for guiding Large Language Models (LLMs) through complex reasoning processes. Our dataset addresses this gap by incorporating comprehensive thought chains, thereby compensating for the limitations of existing geometric datasets.

\textbf{Geometric Reasoning in Multimodal Models.}
The development of multimodal models for geometric reasoning has witnessed substantial progress across multiple dimensions. Early approaches established foundational techniques through targeted data augmentation and architectural innovations: G-LLaVA ~\cite{gao2023gllava} demonstrated effective training data expansion, DFE-GPS ~\cite{lu2021inter} integrated formal geometric languages, and GeoCoder ~\cite{chen2021geoqa} explored modular code generation with retrieval mechanisms. Recent advances have achieved notable breakthroughs: Geo-LLaVA ~\cite{xu2024geo} extended capabilities to three-dimensional scenarios, while AlphaGeometry ~\cite{trinh2024alphageometry} combined neural language models with symbolic deduction engines for remarkable performance. Concurrently, the emergence of Chain-of-Thought prompting ~\cite{wei2022chain} has inspired multimodal extensions including Visual-CoT ~\cite{shao2024visual}and R-CoT ~\cite{deng2024r}, with recent reasoning models like DeepSeek-R1 ~\cite{guo2025deepseek} further demonstrating the potential of extended reasoning processes.

While these approaches have advanced the field significantly, they often emphasize either geometric understanding or reasoning generation as primary focuses, with limited integration of comprehensive thought processes specifically tailored for geometric problem solving. Most existing methods provide final solutions or basic reasoning steps but lack detailed intermediate reasoning that mirrors human problem-solving strategies in geometric contexts. Our GeoThought MLLM contributes to this area by incorporating systematic Chain-of-Thought processes designed specifically for geometric reasoning, enabling models to generate detailed hypothesis formation, step-by-step decomposition, and verification processes while maintaining effective geometric understanding capabilities.
\section{Method}
\label{sec:method}

\begin{figure*}[htbp] 
  \centering 
  \includegraphics[width=1\textwidth]{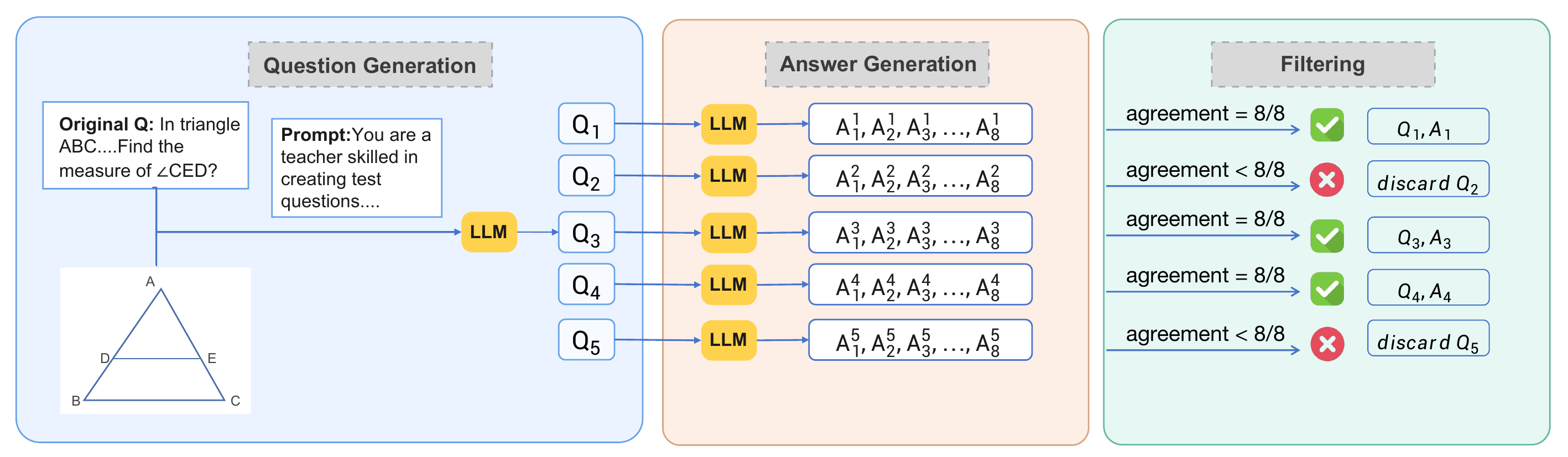} 
  \caption{Our data generation workflow begins by inputting question-image pairs with engineered prompts to produce five novel questions per instance. Each generated question then undergoes eight independent reasoning path generations with corresponding answers. We statistically analyze the answer distributions, retaining only high-confidence QA pairs where a single answer achieves $ \frac{8}{8}$.} 
\label{generation workflow}
\end{figure*}

\subsection{Datasets collection}

Despite existing efforts to address multi-modal geometry problems ~\cite{chen2021geoqa,chen2022unigeo,gao2023gllava}, current geometric datasets exhibit significant limitations: absence of reasoning chains, insufficient scale, and lack of diversity. This poses substantial challenges for MLLMs in advancing geometric reasoning capabilities.
To address the limitations mentioned above in existing datasets, we introduce the GeoThoughts dataset: the construction of the GeoThought-6K dataset and the subsequent expansion to create the GeoThought-Augmented-10K dataset.

\textbf{Geo-Thought-6K construction.}
The Geo-Thought-6K dataset was constructed by utilizing questions and images from the GEOQA\_R1V\_Train\_8K~\cite{leonardpku_geoqa_r1v_train_8k_2024} dataset as inputs to query the Doubao-1.5-thinking-vision-pro model ~\cite{seed2025seed1} for generating thinking processes and answers, using the prompt ``Output the thinking process in \verb|<think>| and final answer in \verb|<answer>| \verb|</answer>| tags.'' This process initially generated 8,031 question-answer pairs with Chain-of-Thought (CoT) reasoning. Subsequently, we applied a rejection sampling approach to filter out samples where the model's answer differed from the ground truth, which ultimately resulted in 6,243 high-quality samples that constitute the final Geo-Thought-6K dataset.

\textbf{Geo-Thought-Augmented-10K expansion.}
The Geo-Thought-Augmented-10K dataset was created through a two-stage process:
\textbf{Step1: Question generation.} We employed data augmentation techniques to diversify the question set using the GEOQA\_R1V\_Train\_8K~\cite{leonardpku_geoqa_r1v_train_8k_2024} dataset as seed data. Through carefully engineered prompts, we instructed the Doubao-1.5-thinking-vision-pro model to generate five additional questions per input instance. The specific prompts used are shown in Table~\ref{table:question-prompt}, with representative examples provided in Table~\ref{table:Example of Generated Question}. \textbf{Step2: Answer and filtering.} To ensure the quality of these newly generated questions, we implemented a post-hoc filtering methodology inspired by ~\cite{wang2022self}. For each generated question, we queried the Doubao-1.5-thinking-vision-pro model eight times, obtaining eight responses. We then extracted answers from each response and computed their frequency distribution, as detailed in Table~\ref{table:answer-frequency}. 
Through manual evaluation, we classified questions as high-confidence when a single answer achieved a probability of 100\% (workflow see Figure~\ref{generation workflow}), ultimately yielding 4,591 additional high-quality samples through this process. 
The final Geo-Thought-Augmented-10K dataset was formed by combining these 4,591 newly generated samples with the original 6,243 samples from Geo-Thought-6K, resulting in a comprehensive collection of 10,834 question-answer pairs. Additional design choice on data generation are provided in the Appendix ~\ref{sec:appendix-pipeline}.

\begin{table}[ht!]
\centering
\renewcommand{\arraystretch}{1.2}
\begin{tabular}{|p{0.45\textwidth}|}
\hline
{\cellcolor[rgb]{0.9,0.9,0.9}}\textbf{Question Prompt} \\
\hline
{\cellcolor[rgb]{0.95,0.97,1.0}}
You are a teacher skilled in creating test questions. Now, an image and a question are provided. Based on the image and the given question, please create 5 additional questions. \textbf{\textcolor{blue!70}{Requirements:}}
\begin{enumerate}
    \item The \textcolor{red}{\textbf{conditions}} and \textcolor{red}{\textbf{problem statements}} (what is asked) must vary.
    \item Our goal is to \textcolor{blue}{\textbf{diversify}} into more varied questions based on the provided image and original question, to support fine-tuning models (we want the model to encounter more questions and more expressions).
    \item The generated questions must \textcolor{green}{\textbf{match the image}}.
    \item Generate \textcolor{purple}{\textbf{short-answer questions}} only; do not generate proof questions.
    \item The questions must be written in \textcolor{orange}{\textbf{English}}.
\end{enumerate}
The image is shown above, and the original question is:
\colorbox{yellow!20}{\texttt{\{QUESTION\}}}
\\
Please output the 5 questions in \textcolor{purple}{\textbf{JSON format}} as a list of objects. Each object should have a key "question" with the question content as its value. Format: \colorbox{blue!10}{\{[{"question": "..."}, {"question": "..."}, ...]\}} \\
\hline
\end{tabular}
\caption{Question Prompt}
\label{table:question-prompt}
\end{table}

\begin{table}[ht!]
\centering
\renewcommand{\arraystretch}{1.2}
\begin{tabular}{|p{0.45\textwidth}|}
\hline
{\cellcolor[rgb]{0.9,0.9,0.9}}\textbf{Input} \\
\hline
{\cellcolor[rgb]{0.95,0.97,1.0}}
\begin{minipage}[t]{\linewidth}
\textbf{\textcolor{blue!70}{Question input:}} \\
As shown in the figure, in triangle ABC, \textcolor{red}{$\angle A = 80.0^\circ$}, \textcolor{red}{$\angle B = 60.0^\circ$}, and \textcolor{blue!70}{DE $\parallel$ BC}. Find the measure of \textcolor{purple}{$\angle CED$} \\
\vspace{10pt} 
\textbf{\textcolor{orange!70}{Image input:}} \\
\includegraphics[width=0.4\linewidth]{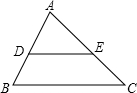}
\vspace{10pt} 
\end{minipage} \\
\hline
{\cellcolor[rgb]{0.9,0.9,0.9}}\textbf{Generated Questions} \\
\hline
{\cellcolor[rgb]{0.98,0.95,1.0}}
\begin{minipage}[t]{\linewidth}
\begin{enumerate}
    \item As shown in the figure, in triangle ABC, \textcolor{blue!70}{DE is parallel to BC}. Given that \textcolor{red}{$\angle B = 60^\circ$}, find the measure of \textcolor{purple}{$\angle ADE$}
   
    \item As shown in the figure, in triangle ABC, \textcolor{blue!70}{DE $\parallel$ BC} with \textcolor{teal}{AD:DB = 1:2}. Find the ratio \textcolor{purple}{DE:BC}
   
    \item As shown in the figure, in triangle ABC, \textcolor{red}{$\angle A = 80^\circ$}, \textcolor{red}{$\angle B = 60^\circ$}, \textcolor{blue!70}{DE $\parallel$ BC}. Determine the degree measure of \textcolor{purple}{$\angle AED$}
   
    \item As shown in the figure, in triangle ABC, \textcolor{blue!70}{DE is the midline} (where D and E are midpoints of AB and AC respectively) and \textcolor{blue!70}{DE $\parallel$ BC}. Calculate the ratio of the \textcolor{orange!70}{perimeter} of $\triangle ADE$ to that of $\triangle ABC$
   
    \item As shown in the figure, in triangle ABC, \textcolor{blue!70}{DE $\parallel$ BC}, \textcolor{teal}{AD = 3}, \textcolor{teal}{AB = 9}. Find the ratio of the \textcolor{orange!70}{area} of $\triangle ADE$ to the area of $\triangle ABC$
\end{enumerate}
\end{minipage} \\
\hline
\end{tabular}
\caption{Example of Generated Questions} 
\label{table:Example of Generated Question}
\end{table}


\begin{table}[ht!]
  \centering
  \footnotesize
  \setlength{\tabcolsep}{3.5pt}
  \begin{tabular}{*{9}{c}} 
  \toprule
  \multicolumn{9}{c}{\textbf{Answer Frequency Distribution}} \\
  \midrule
  \textbf{Consistency} & \textbf{8} & \textbf{7} & \textbf{6} & \textbf{5} & \textbf{4} & \textbf{3} & \textbf{2} & \textbf{1} \\
  \midrule
  \textbf{Percentage (\%)} & 58.4 & 10.2 & 7.5 & 7.4 & 5.9 & 3.2 & 1.5 & 0.6 \\
  \bottomrule
  \end{tabular}
  \caption{Answer Frequency Distribution}
  \label{table:answer-frequency}
\end{table}

\textbf{Data Statistics.}
The Geo-Thought-6K dataset contains answers with an average length of 781 words and reasoning chains averaging 1299 words. Building upon this foundation, the augmented Geo-Thought-10K dataset maintains a comparable answer length of 774 words, while its reasoning chains average 1218 words. To further demonstrate the diversity and comprehensiveness of our dataset, we conducted a statistical analysis of the question types in Geo-Thought-10K. As shown in Appendix ~\ref{sec:question-type-distribution-Geo-Thought-10K}, the dataset covers a wide variety of geometric problem types, including angle calculation, length measurement, area computation, similarity assessment, coordinate geometry, among other geometric concepts. 
This diverse distribution across multiple geometric categories suggests that our dataset encompasses a broad spectrum of geometric reasoning tasks, ensuring comprehensive coverage of different problem types and enhancing its representativeness for training geometric reasoning models.

\subsection{Model Training}

We performed SFT training on the Qwen2.5-VL~\cite{bai2025qwen2} and Intern3VL~\cite{zhu2025internvl3} series. The data format follows the schema established in DeepSeek-R1 ~\cite{guo2025deepseek}, where we encapsulate the reasoning process within \texttt{<think>} tags and the final response within \texttt{<answer>} tags.

\begin{itemize}
    \item \textbf{SFT (Supervised Fine-Tuning)}: 
    The objective function minimizes the negative log-likelihood:
    \[
    \mathcal{L}_{\text{SFT}}(\theta) = -\mathbb{E}_{(x,y) \sim \mathcal{D}_{\text{SFT}}} \left[ \sum_{t=1}^{T} \log \pi_{\theta}(y_t | x, y_{<t}) \right]
    \]
    where $\theta$ are model parameters, $(x,y)$ are input-output pairs from $\mathcal{D}_{\text{SFT}}$, and $T$ is the target sequence length.
\end{itemize}
\section{Experiments}
\label{sec:experiment}
\subsection{Dasets and Metrics}
\noindent \textbf{Training datasets.} We primarily utilized our constructed Geo-Thought-6K and Geo-Thought-10K datasets for training, with the multimodal-open-r1-8k-verified dataset~\cite{lmms_lab_multimodal_open_r1_8k_verified_2024} serving as a control group for comparison. All input images were uniformly resized to a resolution of 336 × 336 pixels.

\noindent \textbf{Testing datasets.} We use GeoQA~\cite{chen2021geoqa} and Geometry3K~\cite{lu2021inter} as our testing datasets, with Geometry3K serving as the out-of-domain evaluation set. GeoQA~\cite{chen2021geoqa} contains 754 test cases, while Geometry3K~\cite{lu2021inter} includes 601 examples.

\noindent \textbf{Mertrcs.} We adopt accuracy as the evaluation metric, ``Acc'' stands for accuracy.) and exclusively focus on reporting Top-1 accuracy. Our evaluation pipeline employs match verify's Python interface to extract per-problem predictions from model outputs, scoring matches with ground truth as correct and mismatches as errors.

\begin{figure*}[!ht] 
    \centering 
    \includegraphics[width=0.98\linewidth]{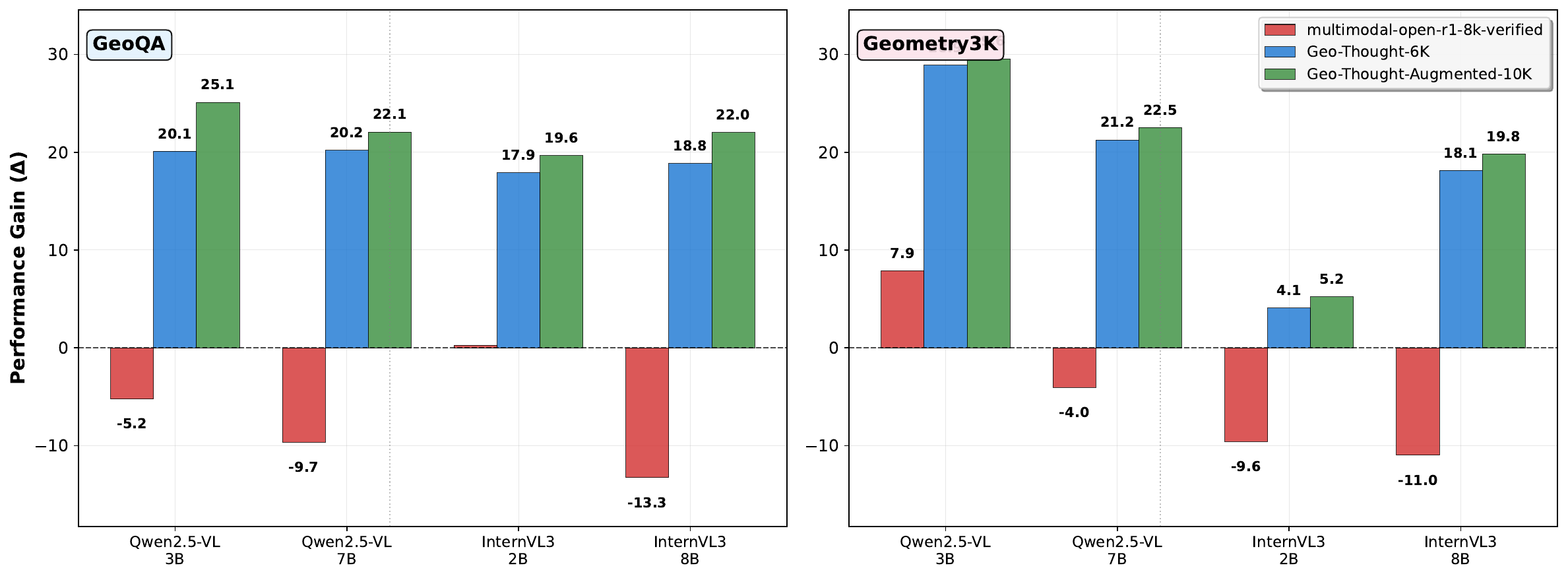} 
    \caption{
        Performance comparison of different models on GeoQA and Geometry3K datasets.
        The bar charts show the performance gains ($\Delta$) achieved by models from the Qwen2.5-VL and Intern3VL series when fine-tuned with different datasets (multimodal-open-r1-8k-verified, Geo-Thought-6K, and Geo-Thought-Augmented-10K) compared to their base versions.
        Results demonstrate that the improvements brought by the Geo-Thought-6K and Geo-Thought-Augmented-10K datasets are particularly significant.
    }
    \label{fig:model_performance_gain} 
\end{figure*}

\subsection{Experiments Setup}

We fine-tune models using AdamW ($\beta_1=0.9$, $\beta_2=0.95$, $\epsilon=1e\text{-}8$, weight-decay$=0.1$) for 3 epochs with an effective global batch size of 256 sequences. The initial learning rate is $1e\text{-}5$ with a linear warmup of 5\% of the total training steps. We apply gradient clipping at 1.0, \texttt{bfloat16} mixed precision, and activation checkpointing. Inputs are truncated to 4096 tokens and targets to 1024 tokens. Training runs on $8 \times$ A100-80GB GPUs under PyTorch 2.0.1 with seed 42. At inference, we use the \texttt{vllm} framework with greedy decoding (temperature $0.0$) and a maximum generation length of 2048 tokens.
\subsection{Main Results}

\noindent \textbf{In-Domain Performance on GeoQA}. Our Geo-Thought datasets demonstrate substantial effectiveness on the in-domain GeoQA benchmark (see Figure~\ref{fig:model_performance_gain}). First, our reasoning dataset, Geo-Thought-6K, demonstrates clear advantages over the existing reasoning dataset, \textit{multimodal-open-r1-8k-verified}. When training the Qwen2.5-VL-3B on \textit{multimodal-open-r1-8k-verified}, we observe a decline in performance, indicating that this dataset may not effectively support model reasoning. In contrast, training on Geo-Thought-6K yields substantial improvements, with gains of +20.10\% on Qwen2.5-VL-3B, +20.20\% on Qwen2.5-VL-7B, +17.91\% on InternVL3-2B, and +18.84\% on InternVL3-8B (see Table~\ref{tab:compact_performance}). As shown in Appendix ~\ref{sec:cot-comparison}, our dataset includes richer thinking and reflection steps, which drive these improvements. Second, incorporating our augmented dataset further boosts performance by an additional +25.10\% on Qwen2.5-VL-3B, +22.06\% on Qwen2.5-VL-7B, +19.65\% on InternVL3-2B, and +22.02\% on InternVL3-8B, respectively. By combining the reasoning and augmentation techniques, our best open-source model, InternVL3-8B, achieves 73.21\%, reaching approximately 94\% of the performance of the closed-source Doubao-1.5-thinking-vision-pro~\cite{seed2025seed1} (77.73\%), highlighting the strong competitiveness of open-source models for complex geometric reasoning.

\noindent \textbf{Out-of-Domain Generalization on Geometry3K}. 
The strong performance of our models extends to the out-of-domain Geometry3K benchmark, highlighting the superior generalization capability gained through our data construction (see Figure~\ref{fig:model_performance_gain}). As illustrated in Table~\ref{tab:compact_performance}, fine-tuning with our Geo-Thought-6K dataset yields substantial cross-domain improvements over base models: +28.91\% for Qwen2.5-VL-3B, +21.25\% for Qwen2.5-VL-7B, +4.07\% for InternVL3-2B, and +18.14\% for InternVL3-8B. The augmented Geo-Thought-Augmented-10K dataset further enhances these gains, achieving improvements of +29.55\%, +22.52\%, +5.23\%, and +19.81\% respectively. Notably, while the existing multimodal-open-r1-8k-verified dataset shows inconsistent effects (even causing performance degradation of -4.05\%, -9.58\%, and -10.98\% on Qwen2.5-VL-7B, InternVL3-2B and InternVL3-8B models respectively), our datasets demonstrate robust positive transfer. This consistent improvement on a completely unseen benchmark confirms that the reasoning patterns learned from our dataset are not limited to in-domain problems but transfer effectively to diverse geometric reasoning tasks, underscoring the generalizability and robustness of our proposed data construction framework.

\newcommand{\annotate}[3]{%
    #1\raisebox{-0.5ex}{\scriptsize\textcolor{#2}{#3}}%
}

\begin{table}[htbp]
\centering
\fontsize{8.5}{10}\selectfont 
\setlength{\tabcolsep}{4pt}
\renewcommand{\arraystretch}{1.1}
\begin{tabular}{@{}lll@{}}
\toprule
\textbf{Model} & \textbf{GeoQA} & \textbf{Geometry3K} \\
\midrule
\multicolumn{3}{l}{{\cellcolor[rgb]{0.95,0.95,0.95}}\textit{\textbf{Closed-source Models}}} \\
Doubao-1.5-thinking-vision-pro & 77.73 & 77.75 \\
\midrule
\multicolumn{3}{l}{{\cellcolor[rgb]{0.95,0.95,0.95}}\textit{\textbf{Open-source Models}}} \\
\textit{Qwen2.5-VL Series} & & \\
\quad Qwen2.5-VL-3B & 38.33 & 22.63 \\
\quad Qwen2.5-VL-7B & 46.15 & 35.77 \\
\quad Qwen2.5-VL-72B & 55.50 & 50.50 \\
\quad QVQ-72B-Preview & 43.10 & 46.17 \\
\addlinespace[2pt]
\textit{InternVL3 Series} & & \\
\quad InternVL3-2B & 32.36 & 35.20 \\
\quad InternVL3-8B & 51.19 & 45.09 \\
\addlinespace[2pt]
\textit{Other Models} & & \\
\quad G-LLaVA-7B & 64.2 & 28.79 \\
\quad G-LLaVA-13B & 64.7 & 27.45 \\
\quad GLM-4.5V & 76.25 & 70.39 \\
\midrule
\multicolumn{3}{l}{{\cellcolor[rgb]{0.95,0.95,0.95}}\textit{\textbf{Finetuned Models (Training Data → Performance)}}} \\
\textit{Qwen2.5-VL-3B} & & \\
\quad multimodal-open-r1-8k-verified & 33.11 \annotate{}{blue}{(-5.22)} & 30.50 \annotate{}{red}{(+7.87)} \\
\quad Geo-Thought-6K & 58.43 \annotate{}{red}{(+20.10)} & 51.54 \annotate{}{red}{(+28.91)} \\
\quad Geo-Thought-Augmented-10K & \textbf{63.43 \annotate{}{red}{(+25.10)}} & \textbf{52.18 \annotate{}{red}{(+29.55)}} \\
\addlinespace[2pt]
\textit{Qwen2.5-VL-7B} & & \\
\quad multimodal-open-r1-8k-verified & 36.50 \annotate{}{blue}{(-9.65)} & 31.72 \annotate{}{blue}{(-4.05)} \\
\quad Geo-Thought-6K & 66.35 \annotate{}{red}{(+20.20)} & 57.02 \annotate{}{red}{(+21.25)} \\
\quad Geo-Thought-Augmented-10K & \textbf{68.21 \annotate{}{red}{(+22.06)}} & \textbf{58.29 \annotate{}{red}{(+22.52)}} \\
\addlinespace[2pt]
\textit{InternVL3-2B} & & \\
\quad multimodal-open-r1-8k-verified & 32.63 \annotate{}{red}{(+0.27)} & 25.62 \annotate{}{blue}{(-9.58)} \\
\quad Geo-Thought-6K & 50.27 \annotate{}{red}{(+17.91)} & 39.27 \annotate{}{red}{(+4.07)} \\
\quad Geo-Thought-Augmented-10K & \textbf{52.01 \annotate{}{red}{(+19.65)}} & \textbf{40.43 \annotate{}{red}{(+5.23)}} \\
\addlinespace[2pt]
\textit{InternVL3-8B} & & \\
\quad multimodal-open-r1-8k-verified & 37.93 \annotate{}{blue}{(-13.26)} & 34.11 \annotate{}{blue}{(-10.98)} \\
\quad Geo-Thought-6K & 70.03 \annotate{}{red}{(+18.84)} & 63.23 \annotate{}{red}{(+18.14)} \\
\quad Geo-Thought-Augmented-10K & \textbf{73.21 \annotate{}{red}{(+22.02)}} & \textbf{64.90 \annotate{}{red}{(+19.81)}} \\
\bottomrule
\end{tabular}
\caption{Model Performance Comparison and Improvement Results}
\label{tab:compact_performance}
\end{table}

\begin{table}[h!]
\centering
\renewcommand{\arraystretch}{1.2}
\begin{tabular}{|p{0.45\textwidth}|}
\hline
{\cellcolor[rgb]{0.9,0.9,0.9}}\textbf{Input} \\
\hline
{\cellcolor[rgb]{0.95,0.97,1.0}}
\begin{minipage}[t]{\linewidth}
\textbf{Question input:} \\
As shown in the figure, in triangle ABC, it is known that angle A = 80.0, angle B = 60.0, DE parallel to BC, then the size of angle CED is ().
\vspace{1pt}
\textbf{Image input:} \\
\includegraphics[width=0.25\linewidth]{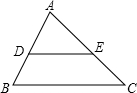} 
\vspace{1pt}
\end{minipage} \\
\hline
{\cellcolor[rgb]{0.9,0.9,0.9}}\textbf{Reasoning Process From Model} \\
\hline
{\cellcolor[rgb]{0.98,0.95,1.0}}
\begin{minipage}[t]{\linewidth}
\par
Got it, let's see. The problem is about a triangle ABC with DE parallel to BC. We need to find the measure of angle CED.\textcolor{blue!60}{\textless{}omitted\textgreater{}}
Now, DE is parallel to BC. So DE $||$ BC, \textcolor{red}{which means that angle CED and angle C are corresponding angles}. right? Because DE is parallel to BC, and EC is a transversal. Corresponding angles are equal when two parallel lines are cut by a transversal. So angle CED should be equal to angle C. 
Therefore, angle CED is 40 degrees.
\par
\vspace{1pt}
\end{minipage} \\
\hline
\end{tabular}
\caption{The multimodal model incorrectly judged the positional relationship of the elements in the geometric figure: angle CED and angle C should be consecutive interior angles, but the model misidentified them as corresponding angles.}
\label{tab:error_analysis_angle_relationship}
\end{table}

\begin{table}[h!]
\centering
\begin{tabular}{|p{0.45\textwidth}|}
\hline
{\cellcolor[rgb]{0.9,0.9,0.9}}
\textbf{Input} \\
\hline
\begin{minipage}[t]{\linewidth}
{\cellcolor[rgb]{0.95,0.97,1.0}}
\textbf{Question input:} \\
As shown in the figure, in triangle ABC, it is known that angle A = 80.0, angle B = 60.0, DE parallel to BC, then the size of angle CED is ().

\vspace{1pt}
\textbf{Image input:} \\
\includegraphics[width=0.25\linewidth]{figures/figure_4.png} 
\vspace{1pt}
\end{minipage} \\
\hline
{\cellcolor[rgb]{0.9,0.9,0.9}}
\textbf{Reasoning Process From Model} \\
\hline
\begin{minipage}[t]{\linewidth}
\vspace{0pt}
\par
{\cellcolor[rgb]{0.98,0.95,1.0}}
Got it, let's see. The problem is about a triangle ABC with DE parallel to BC. We need to find the measure of angle CED.
\textcolor{blue!60}{\textless{}omitted\textgreater{}}
Now, DE is parallel to BC. So DE $||$ BC, \textcolor{red}{which means that angle CED and angle C are corresponding angles}. right? \textcolor{blue!60}{\textless{}omitted\textgreater{}}
\par
\vspace{1pt}
\end{minipage} \\
\hline
{\cellcolor[rgb]{0.9,0.9,0.9}}
\textbf{Follow-up Questioning and Model’s Response} \\
\hline
\begin{minipage}[t]{\linewidth}
\vspace{0pt}
{\cellcolor[rgb]{0.98,0.95,1.0}}
\textbf{Q: What are corresponding angles? Why are angle CED and angle C considered corresponding angles?}
\par
\vspace{0.5em}
\textbf{Model's Response:} \\
According to the definition of corresponding angles: \\
When two lines are intersected by a third line (the transversal), the angles that lie on the same side of the transversal and in the same direction with respect to the two intersected lines are called \textcolor{red}{corresponding angles.} \\
In this problem: \\
- The transversal EC intersects the two lines ED and BC; \\
- angle CED and angle C are both on the same side of the transversal EC; \\
- Meanwhile, \textcolor{red}{angle CED is ``above'' the intersected line ED}, and angle C is ``above'' the intersected line BC (they are in the same relative direction).
\par
\vspace{1pt}
\end{minipage} \\
\hline
\end{tabular}
\caption{The multimodal model incorrectly judged the positional relationship of the elements in the geometric figure: angle CED and angle C should be consecutive interior angles, but the model misidentified them as corresponding angles. The model's follow-up explanation further demonstrates its spatial misjudgment by incorrectly locating angle CED ``above” line ED.}
\label{table:spatial-misjudgment-example- cause}
\end{table}

\begin{table}[h!]
\centering
\begin{tabular}{|p{0.45\textwidth}|}
\hline
{\cellcolor[rgb]{0.9,0.9,0.9}}
\textbf{Input} \\
\hline
\begin{minipage}[t]{\linewidth}
{\cellcolor[rgb]{0.95,0.97,1.0}}
\textbf{Text input:} \\
As shown in the figure, in triangle ABC, it is known that angle A = 80.0, angle B = 60.0, DE parallelBC, then the size of angle CED is ()

\verb|<think>| Got it, let's see. The problem is about a triangle ABC with DE parallel to BC. \textcolor{blue!60}{\textless{}omitted\textgreater{}}That's straightforward. Now, DE is parallel to BC, \textcolor{red}{which means that angle CED and angle C are consecutive interior angles, }( \textbf{a corrected but truncated ``Thinking” process})

\vspace{1pt}
\textbf{Image input:} \\
\includegraphics[width=0.25\linewidth]{figures/figure_4.png} 
\vspace{1pt}
\end{minipage} \\
\hline
{\cellcolor[rgb]{0.9,0.9,0.9}}
\textbf{Reasoning Process From Model} \\
\hline
\begin{minipage}[t]{\linewidth}
\vspace{0pt}
\par
{\cellcolor[rgb]{0.98,0.95,1.0}}
First, let's find $\angle C$ in triangle $ABC$. Since $\angle A = 80^\circ$ and $\angle B = 60^\circ$, we use the fact that the sum of angles in a triangle is $180^\circ$. So:
$
\angle C = 180^\circ - \angle A - \angle B = 180^\circ - 80^\circ - 60^\circ = 40^\circ
$
Now, $DE \parallel BC$. When two lines are parallel, consecutive interior angles are supplementary (they add up to $180^\circ$). \textcolor{red}{Here, angle CED and angle C are consecutive interior angles} because $DE \parallel BC$ and $AC$ is a transversal. 
\textcolor{blue!60}{\textless{}omitted\textgreater{}}Thus, the size of $\angle CED$ is $\boldsymbol{140^\circ}$(\textbf{the answer is right}).
\end{minipage} \\
\hline
\end{tabular}
\caption{If those mistakes have been corrected in the CoT, the model can eventually answer correctly. }
\label{tab:Example of model reasoning with corrected process}
\end{table}

\subsection{Analysis}

In the previous section, we demonstrated the effectiveness of our CoT and augmented datasets. In this section, we analyze the failure cases of our best-performing model to identify key challenges and provide insights that can guide future research improvements. We observed that $26.79\%$ of the answers in the GeoQA test set remain incorrect after SFT training. We have analyzed these erroneous cases and identified several common issues.

\noindent
\textbf{Observation: Model makes mistakes in visual mathematical concepts.} We found that 87.5\% of these errors stem from the misjudgment of spatial relationships in geometric diagrams by multimodal models during the reasoning process. 
For instance, as detailed in Table ~\ref{tab:error_analysis_angle_relationship}, models misidentify consecutive interior angles as corresponding angles. To further investigate this issue, we prompted the model to explain the concept of ``corresponding angles'' and specifically asked it to justify why angle CED and angle C should be considered as such. 
Although the model provided a correct definition of corresponding angles, it misjudged the spatial position of angle CED.
The angle should lie below line ED, but the model incorrectly located it above line ED (as shown in Table ~\ref{table:spatial-misjudgment-example- cause}. This positional error led the model to misclassify consecutive interior angles as corresponding angles, resulting in an incorrect conclusion.

\vspace{0.5em}

\noindent
\textbf{CoT corrections: if those mistakes have been corrected in the CoT, the model can eventually answer correctly. } Following this, we hypothesized whether the model would produce correct answers if errors in its judgment of spatial relationships during reasoning were corrected. We therefore conducted a small-scale test using the fine-tuned Qwen2.5-VL-7B model. By manually correcting errors in the model’s “Thinking” process and prompting it to continue generating the response, 
we observed that the model was able to produce correct answers (as shown in Table ~\ref{tab:Example of model reasoning with corrected process}).
Concretely, we provided the model with the image, the question, and a corrected but truncated version of the ``Thinking” process, and then instructed the model to complete the reasoning.
\noindent

\noindent
\textbf{Other potential solutions.} From the previous tests, the model's errors in judging spatial relationships among geometric elements during reasoning are mainly due to ``inaccurate visual perception”. Some researchers attribute this phenomenon to \textit{visual forgetting} ~\cite{sun2025mitigating}. Visual forgetting refers to the tendency of MLLMs to gradually reduce their attention to visual inputs during long-chain reasoning, relying more on previously generated text for subsequent inference. Currently, there are two main approaches to mitigate visual forgetting. One method involves selecting multiple self-reflection points during reasoning to reinject visual information, allowing the model to continue inference based on the newly injected visual cues ~\cite{sun2025mitigating}. Another approach employs an agent framework that facilitates multi-turn interaction and tool invocation, re-inserting target regions of the image into the dialogue history ~\cite{zheng2025deepeyes}. Both methods aim to enable the multimodal model to ``look at the image again” during reasoning, achieving the goal of interleaving ``thinking” and ``seeing”. These strategies represent promising directions for future research and are key focuses of our ongoing work.
\vspace{0.5em}

\subsection{Ablation Study}
In this section, we conducted extensive ablation experiments, demonstrating that our chain-of-thought methodology effectively enhances multimodal language models.

\noindent\textbf{Training with and without CoT.}  We performed comparative experiments between datasets with CoT annotations and those without. This comparison revealed significant performance improvements when explicit reasoning pathways were included, confirming that structured reasoning steps substantially boost model comprehension and output quality in multimodal tasks, see Table ~\ref{tab:Effect of data COT on downstream accuracy}.


\begin{table}[h!]
\centering
\footnotesize
\setlength{\tabcolsep}{8pt}
\begin{tabular}{*{3}{c}} 
\toprule
\textbf{Model} & \textbf{acc w/o CoT (\%)} & \textbf{acc w/ CoT (\%)} \\
\midrule
Qwen2.5-VL-3B & 28.92 & 67.1 \\
\bottomrule
\end{tabular}
\caption{Effect of data CoT on downstream accuracy.}
\label{tab:Effect of data COT on downstream accuracy}
\end{table}

\noindent\textbf{Training with filtering and without filtering data.} We conducted an ablation study on the SFT performance of Qwen2.5-VL-3B using both unfiltered and filtered data. Specifically, we generated 7,445 new samples from a subset of the original training set. Among these, 4,591 samples achieved unanimous agreement across all eight rounds of voting. These high-consensus samples were merged with the original 6,243 samples to form the filtered dataset. In contrast, the unfiltered dataset was constructed by combining all 7,445 generated samples (without any filtering) with the original 6,243 samples. The experimental results demonstrate that incorporating the unfiltered data led to only a marginal performance improvement +0.03\%, whereas using the filtered data resulted in a substantial gain of +5.00\% in accuracy, underscoring the critical role of data quality over mere quantity, see Table ~\ref{tab:filtering_ablation}.

\begin{table}[h]
\centering
\small 
\setlength{\tabcolsep}{4pt} 
\begin{tabular}{lccc}
\toprule
\textbf{Training Data} & \textbf{Sample} & \textbf{Accuracy} & \textbf{$\Delta$} \\
\textbf{Configuration} & \textbf{Size} & \textbf{(\%)} & \textbf{(\%)} \\
\midrule
Geo-Thought-6K (Original) & 6,243 & 58.43 & -- \\
+ Unfiltered Generated Data & 13,688 & 58.46 & +0.03 \\
+ \textbf{Filtered Generated Data} & \textbf{10,834} & \textbf{63.43} & \textbf{+5.00} \\
\bottomrule
\end{tabular}
\caption{
    Effectiveness of the answer filtering strategy. Incorporating unfiltered data brings negligible performance gains +0.03\%, while adding the filtered data leads to a substantial improvement of +5.00\%, conclusively validating the effectiveness of our proposed filtering approach.
}
\label{tab:filtering_ablation}
\end{table}

\noindent\textbf{Training with RL.} We also experimented with different training methodologies: one following a GRPO~\cite{shao2024deepseekmath} paradigm and another employing an SFT+GRPO approach. In the GRPO setting, we used the Geo-Thought-6K dataset. The SFT+GRPO method consisted of two stages: the SFT phase also employed the Geo-Thought-6K dataset, while the RL phase utilized 1,788 questions that were filtered out during rejection sampling from an original set of 8,031 samples. Specifically, these questions corresponded to problems that Doubao-1.5-thinking-vision-pro answered incorrectly in a single sampling attempt. The results indicated that neither RL-based approach outperformed the use of SFT alone with the Geo-Thought-6K dataset, see Table ~\ref{tab:model_performance_comparison-v3}. We attribute this outcome to two main factors. First, both the quality and quantity of the data used during the reinforcement learning phase appear to have adversely influenced the results~\cite{li2025limr}. Additionally, in the SFT+GRPO paradigm, the initial SFT stage may have overly constrained the exploration space available for subsequent GRPO optimization~\cite{chen2025sft}.

\begin{table}[h]
\centering
\small 
\begin{tabular}{lcl}
\toprule
\textbf{Model} & \textbf{Training Strategy} & \textbf{Performance} \\
\midrule
Qwen2.5-VL-7B & SFT Only & 66.35 \\
Qwen2.5-VL-7B & GRPO Only & 56.62 \annotate{}{blue}{(-9.73)} \\
Qwen2.5-VL-7B & SFT + GRPO & 65.85 \annotate{}{blue}{(-0.50)} \\
\midrule
Qwen2.5-VL-3B & SFT Only & 58.43 \\
Qwen2.5-VL-3B & GRPO Only & 52.43 \annotate{}{blue}{(-6.00)} \\
Qwen2.5-VL-3B & SFT + GRPO & 57.24 \annotate{}{blue}{(-1.19)} \\
\bottomrule
\end{tabular}
\caption{Model Performance Comparison with Different Training Strategies}
\label{tab:model_performance_comparison-v3}
\end{table}
\section{Conlustion}
\label{sec:conclusion}

In this paper, we tackle key challenges in geometric reasoning for multimodal large language models by introducing GeoThoughts, a high-quality chain-of-thought dataset comprising Geo-Thought-6K and its augmented version, Geo-Thought-Augmented-10K. Each sample embodies high-quality reasoning patterns such as problem decomposition, reflective verification, and step backtracking. Based on this dataset, we developed GeoThought-MLLM, a compact model capable of generating structured reasoning processes during problem-solving. Experimental results show that our model significantly outperforms existing benchmarks on both in-domain and out-of-domain geometric tasks, achieving performance comparable to state-of-the-art closed-source models, confirming the efficacy of explicit Chain-of-Thought training.

Furthermore, analysis of erroneous cases from experiments reveals that most failures stem from misjudgments of spatial geometric relationships within the reasoning chain. Importantly, we demonstrate that targeted corrections to the reasoning process enable the model to successfully reach correct solutions. Therefore, enabling models to accurately judge spatial relationships among geometric elements during reasoning will be a key direction for our future research.

We openly release the GeoThoughts dataset and model to support further research in multimodal reasoning, with the goal of promoting advances in geometrically-grounded AI systems.
{
\bibliographystyle{ieeenat_fullname}
    \bibliography{main}
}
\begin{appendix}
\clearpage
\section{appendix}
\label{sec:appendix}

\subsection{Data Generation Pipeline Details}
\label{sec:appendix-pipeline}
We outline the key methodological choices made in constructing the GeoThoughts dataset, including choice of data format and teacher model selection.

\textbf{Choice Of Data Format.} Publicly available geometric reasoning datasets typically come in two formats: multiple-choice questions, such as the multimodal-open-r1-8k-verified~\cite{lmms_lab_multimodal_open_r1_8k_verified_2024}, and open-ended questions, like those in GEOQA\_R1V\_Train\_8K~\cite{leonardpku_geoqa_r1v_train_8k_2024}. However, we found that when using multiple-choice questions to generate chains of thought, the model often failed to perform valid reasoning and instead randomly selected an answer from the options. Consequently, training multimodal models with such chain-of-thought data, which contains substantial random noise, can severely impair their robust reasoning capabilities. Therefore, to address this issue, we sampled both questions and images from the open-ended GEOQA\_R1V\_Train\_8K dataset to elicit high-quality chain-of-thought reasoning processes.

\textbf{Teacher Model Selection.}
    We evaluated GLM-Zero-Preview~\cite{zhipuai_glm_zero_preview_2024} and Doubao-1.5-thinking-vision-pro as teacher models for chain-of-thought generation and found that the latter produces reasoning chains of significantly higher quality. Doubao-1.5-thinking-vision-pro is particularly distinguished by two key strengths: its reasoning process closely mirrors human problem-solving strategies, and it performs reflection and verification with high frequency, enabling verification at nearly every step to ensure timely error correction.

At the beginning of its reasoning chain, Doubao-1.5-thinking-vision-pro leverages relevant theorems to uncover hidden geometric conditions as exhaustively as possible, much like humans do when solving problems. For example, upon identifying triangle ABC as isosceles, it immediately invokes the base angles theorem to deduce that angles B and C are equal(see Table ~\ref{tab:Reasoning Process From Doubao-1.5-thinking-vision-pro}). Such insights often prove critical to resolving the problem. In contrast, GLM-Zero-Preview shows limited capability in detecting these hidden conditions during early reasoning stages.

Moreover, Doubao-1.5-thinking-vision-pro demonstrates a frequent reflection mechanism. For instance, after inferring that ‘‘ the median to the base of an isosceles triangle is also the altitude and the angle bisector, ’’ the model pauses to verify the correctness of its reasoning, often performing multiple verification checks within a single reasoning step. While GLM-Zero-Preview does possess some reflective ability, its lower frequency of validation makes it susceptible to accumulating hallucinations without timely correction(see Table ~\ref{tab:Reasoning Process From GLM-Zero-Preview}). Ultimately, Doubao-1.5-thinking-vision-pro further enhances reliability by conducting a comprehensive secondary review of all steps upon completing the chain of thought, a global verification mechanism entirely absent in the GLM-Zero-Preview.

Therefore, we selected Doubao-1.5-thinking-vision-pro as our teacher model.

\begin{table}[h!]
\centering
\small 
\begin{tabular}{|p{0.45\textwidth}|}
\hline
\textbf{Input} \\
\hline
\begin{minipage}[t]{\linewidth}
\textbf{Question input:} \\
In the given figure, triangle ABC is circumscribed about circle O. The external angle BAE's bisector intersects circle O at point D, and point A is the midpoint of arc CD. If the angle ABC is 28°, what is the measure of angle ACB?

\textbf{Choices:} A: 84°, B: 85°, C: 86°, D: 88° \\
\vspace{1pt}
\textbf{Image input:} \\
\includegraphics[width=0.25\linewidth]
{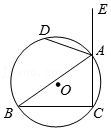}
\vspace{1pt}
\end{minipage} \\
\hline
\textbf{Reasoning Process From GLM-Zero-Preview} \\
\hline
\begin{minipage}[t]{\linewidth}
\vspace{0pt}
\par
Let's tackle this geometry problem step by step. We have triangle \( ABC \) circumscribed about circle \( O \) , meaning the circle \textcolor{blue!60}{\textless{}omitted\textgreater{}}
Multiplying both sides by 2:
$
304^\circ = 28^\circ + C
$
Subtracting $28^\circ$ from both sides:
$
C = 276^\circ
(\textbf{reasoning error})
$
Wait, that can't be right. An angle in a triangle can't be $ 276^\circ $. 
\textcolor{blue!60}{\textless{}omitted\textgreater{}}
Let's try a different approach.
\textcolor{blue!60}{\textless{}omitted\textgreater{}}
So:
$
- C = C \Rightarrow 2C = 0 \Rightarrow C = 0^\circ
(\textbf{reasoning error})$
That can't be right either.
\textcolor{blue!60}{\textless{}omitted\textgreater{}}
After much consideration, it seems that the measure of \( \angle ACB \) is \( 86^\circ \)(\textbf{Failed to deduce, guessed randomly}).

\vspace{1pt}
\end{minipage} \\
\hline
\end{tabular}
\caption{When presented with multiple-choice questions, model that fail to reason correctly may randomly select answer from given options, introducing stochastic noise into the inference process.}
\label{tab:Reasoning Process From GLM-Zero-Preview}
\end{table}

\begin{table}[h!]
\centering
\small 
\begin{tabular}{|p{0.48\textwidth}|}
\hline
\textbf{Reasoning Process From Doubao-1.5-thinking-pro} \\
\hline
\begin{minipage}[t]{0.48\textwidth}
\vspace{6pt}

Got it, let's try to solve this problem step by step. First, let's recall what we know about triangle ABC. It's an isosceles triangle with AB = AC, so that means the base is BC, and the base angles at B and C are equal \textbf{(leverages relevant theorems to uncover hidden geometric conditions )}.
\textcolor{blue!60}{\textless{}omitted\textgreater{}}
Next, AD is the median of side BC. In an isosceles triangle, the median from the vertex (A) to the base (BC) is also the altitude and the angle bisector. So AD is perpendicular to BC? Wait, no, only if it's an equilateral triangle, but here it's isosceles. Wait, actually, in an isosceles triangle, the median from the apex (A) to the base (BC) is also the altitude and the angle bisector \textbf{(verify the correctness of its reasoning step-by-step)}.
\textcolor{blue!60}{\textless{}omitted\textgreater{}}
Wait, let's make sure about the angles. Let's confirm each step \textbf{(a comprehensive secondary review of all steps)}:

\begin{enumerate}
    \item AB = AC, so triangle ABC is isosceles with $\angle$ABC = $\angle$ACB. $\angle$BAC = 100°, so $\angle$ABC + $\angle$ACB = 80°, so each is 40°. Correct.
    \item AD is the median of BC, so D is the midpoint of BC.
\textcolor{blue!60}{\textless{}omitted\textgreater{}}    
\end{enumerate}

\vspace{6pt}
\end{minipage} \\
\hline
\end{tabular}
\caption{Doubao-1.5-thinking-pro demonstrates integrated reasoning that not only leverages relevant theorems to uncover hidden geometric conditions in a manner closely aligned with human problem-solving approaches, but also  performing frequent reflection and verification.}
\label{tab:Reasoning Process From Doubao-1.5-thinking-vision-pro}
\end{table}

\subsection{Question Type Distribution in Geo-Thought-10K Dataset}
\label{sec:question-type-distribution-Geo-Thought-10K}
Table~\ref{table:question-type-distribution-Geo-Thought-10K} shows the distribution of question types in our dataset. Angle problems dominate (63.5\%), reflecting their fundamental role in geometric reasoning, followed by length calculations (18.3\%) and area problems (9.5\%). This distribution aligns with educational curricula where angular relationships form the foundation for understanding complex geometric concepts. The balanced representation across different problem types ensures comprehensive coverage of essential geometric reasoning skills.

\begin{table}[h!]
  \centering
  \footnotesize
  \setlength{\tabcolsep}{2pt}
  \begin{tabular}{*{7}{c}}
  \toprule
  \multicolumn{7}{c}{\textbf{Question Type Distribution in Geo-Thought-10K Dataset}} \\
  \midrule
  \textbf{Type} & 
  \begin{tabular}{@{}c@{}}\textbf{Angle}\end{tabular} & 
  \begin{tabular}{@{}c@{}}\textbf{Length}\end{tabular} & 
  \begin{tabular}{@{}c@{}}\textbf{Area}\end{tabular} & 
  \begin{tabular}{@{}c@{}}\textbf{Similarity}\end{tabular} & 
  \begin{tabular}{@{}c@{}}\textbf{Coordinate}\end{tabular} & 
  \begin{tabular}{@{}c@{}}\textbf{Other}\end{tabular} \\
  \midrule
  \textbf{Percentage (\%)} & 63.5 & 18.3 & 9.5 & 5.0 & 3.4 & 0.2 \\
  \bottomrule
  \end{tabular}
  \caption{Question Type Distribution in Geo-Thought-10K Dataset}
  \label{table:question-type-distribution-Geo-Thought-10K}
\end{table}

\subsection{Comparison of Chain-of-Thought reasoning processes from different datasets}
\label{sec:cot-comparison}
To demonstrate the quality of reasoning in our GeoThought dataset, we compare Chain-of-Thought processes with existing datasets using the same geometric problem. Table~\ref{table:cot-comparison} reveals key differences: while multimodal-open-r1-8k-verified provides direct solutions, our Geo-Thought-6K demonstrates deeper reasoning through explicit self-verification ("Wait, let me check that again"), systematic error checking, and consideration of alternative approaches. This enhanced reasoning process mirrors human problem-solving strategies and provides superior training data for developing robust geometric reasoning capabilities.

\begin{table}[h!]
\centering
\small 
\begin{tabular}{|p{0.48\textwidth}|}
\hline
\textbf{Task} \\
\hline
\begin{minipage}[t]{\linewidth}
Consider triangle ABC, where the midpoints of sides BC and BA are represented by points D and E respectively. If the area of triangle ABC is given as 32, what is the area of triangle DEB?
\vspace{1pt}
\end{minipage} \\
\hline
\textbf{CoT from multimodal-open-r1-8k-verified} \\
\hline
\begin{minipage}[t]{\linewidth}
\vspace{0pt}
Let's start by understanding the relationship between the triangles. Since D and E are midpoints, triangle DEB is similar to triangle ABC, and its area is a fraction of the area of triangle ABC. First, note that triangle DEB is half of triangle ABD because E is the midpoint of BA. Similarly, triangle ABD is half of triangle ABC because D is the midpoint of BC. Therefore, the area of triangle DEB is 0.5 times the area of triangle ABD, and the area of triangle ABD is 0.5 times the area of triangle ABC. So, the area of triangle DEB is 0.5 * 0.5 * area of triangle ABC. Substitute the given area of triangle ABC, which is 32, into the expression: 0.5 * 0.5 * 32 = 8.
\vspace{1pt}
\end{minipage} \\
\hline
\textbf{CoT from Geo-thought-6K} \\
\hline
\begin{minipage}[t]{\linewidth}
\vspace{0pt}
Got it, let's try to solve this problem step by step. First, let's recall some properties of triangles and midpoints. The problem says that D and E are midpoints of BC and BA respectively, and the area of triangle ABC is 32. We need to find the area of triangle DEB.

First, let's remember that the area of a triangle is (base * height)/2. When we have midpoints, they can divide sides into equal parts, which might help in finding the base and height of smaller triangles relative to the original one.

Let's start by looking at triangle ABD. Since D is the midpoint of BC, BD = DC. So, the base BD is half of BC. Now, triangles ABD and ADC share the same height from A to BC. Therefore, their areas should be equal. Because the area of ABC is 32, the area of ABD is half of that, so 32 / 2 = 16. So area of ABD is 16.
Now, let's look at triangle DEB.\textcolor{blue!60}{\textless{}omitted\textgreater{}}
Wait, let me check that again. Let's verify each step.\textcolor{blue!60}{\textless{}omitted\textgreater{}}
Then, E is the midpoint of BA, so BE = BA/2.\textcolor{blue!60}{\textless{}omitted\textgreater{}}


Alternatively, we can think in terms of midsegments or similar triangles, but since E and D are midpoints, DE is a midsegment of triangle ABC, so DE is parallel to AC and DE = (1/2)AC. But maybe that's more complicated. The key is using the midpoints to divide the area into smaller equal parts.

So first, D is the midpoint of BC, so area of ABD is half of ABC: 32 / 2 = 16. Then E is the midpoint of AB, so area of DEB is half of ABD: 16 / 2 = 8. Therefore, the area of triangle DEB is 8.
\vspace{1pt}
\end{minipage} \\
\hline
\end{tabular}
\caption{Comparison of Chain-of-Thought reasoning processes from different datasets: Geo-thought-6K demonstrates stronger reasoning and thinking capabilities through explicit self-verification steps, error checking, and multiple reasoning pathways, while multimodal-open-r1-8k-verified provides a more linear, assumption-based reasoning process without validation mechanisms.}
\label{table:cot-comparison}
\end{table}

\end{appendix}
\end{document}